\documentclass{article}
\usepackage{spconf,amsmath,graphicx}
\usepackage{amssymb}
\usepackage{xcolor}
\usepackage{enumitem}
\setlist{nosep, leftmargin=14pt}

\usepackage{multirow}
\usepackage{hyperref}
\usepackage{xspace}
\newcommand{\mName}{CLEAR\xspace}
\title{CLEAR: Causal Learning Framework for Robust Histopathology Tumor Detection under Out-of-distribution Shifts \thanks{
\textcopyright 2026 IEEE. Personal use of this material is permitted. Permission from IEEE must be obtained for all other uses, including reprinting/republishing this material for advertising or promotional purposes, collecting new collected works for resale or redistribution to servers or lists, or reuse of any copyrighted component of this work in other works.
}}
\name{Kieu-Anh Truong Thi $^{1}$,
\textit{Huy-Hieu Pham $^{2, 3}$, Duc-Trong Le $^{1,*}$ \thanks{* Corresponding author.}}}
\address{
$^{1}${VNU University of Engineering and Technology, Hanoi 100000, Vietnam}\\
$^{2}$College of Engineering and Computer Science, VinUniversity, Hanoi 100000, Vietnam \\
$^{3}${VinUni-Illinois Smart Health Center, VinUniversity,  Hanoi 100000, Vietnam}\\
}

\begin{document}
\maketitle

\begin{abstract} 
Domain shift in histopathology, often caused by differences in acquisition processes or data sources, poses a major challenge to the generalization ability of deep learning models. Existing methods primarily rely on modeling statistical correlations by aligning feature distributions or introducing statistical variation, yet they often overlook causal relationships. In this work, we propose a novel causal learning framework named CLEAR that leverages semantic features while mitigating the impact of confounders. It implements the front-door principle by designing transformation strategies that explicitly incorporate mediators and observed tissue slides. We validate our method on the CAMELYON17 dataset and a private histopathology dataset, demonstrating consistent performance gains across unseen domains. As a result, our approach achieved up to a 7\% improvement in both the CAMELYON17 dataset  and the private histopathology dataset, outperforming existing baselines. These results highlight the potential of causal inference as a powerful tool for addressing domain shift in histopathology image analysis.
\end{abstract}

\section{Introduction}
A key challenge that limits the real-world use of whole slide image (WSI) analysis with machine learning is the out-of-distribution (OOD) problem. This problem occurs when the data at test time differ from the data used for training. In histopathology, scanners, imaging devices, and staining protocols vary across laboratories. These differences cause changes in illumination and color characteristics \cite{rashmi2022breast}. Stacke et al. \cite{stacke2020measuring} showed that such domain shifts significantly reduce model performance on H\&E-stained images.
\begin{figure}
    \centering
    \includegraphics[width=1\linewidth]{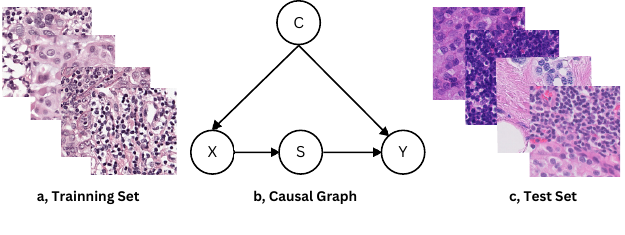}
    \caption{a, c, Example images from training and test datasets. Stain variation can observed between domains.  b, Causal graph represent the causal relationship among the confounder C, the input image X, the mediator S and the label Y}
    \label{fig:causal-graph}
\end{figure}

Recently, several methods have been proposed to reduce domain shift errors. Stain color normalization minimizes stain variation by standardizing the color distribution between training and test images \cite{vahadane2016structure, macenko2009method, reinhard2001color}. Other studies highlight the role of stain color augmentation in improving generalization by exposing the model to diverse image appearances during training \cite{tellez2019quantifying, yamashita2021learning, pohjonen2022augment}. However, most existing methods focus on learning domain-invariant features or aligning color distributions between the source and target domains.

We recognize that histopathology datasets from different domains may contain domain-specific features that influence model predictions. However, the underlying causal relationships represented by semantic features remain consistent across environments. These causal factors provide stable reasoning for prediction. Motivated by this idea, we propose CLEAR, a \underline{C}ausal \underline{LEAR}ning framework for robush histopathology tumor detection under out-of-distribution shifts. It applies the front-door principle by formulating transformation strategies that explicitly account for mediators and observed tissue slides. To evaluate the effectiveness of CLEAR, we conduct experiments on the CAMELYON17 and a private histopathology datasets. Overall, our contributions are summarized as follows:

\begin{figure*}[t!]
    \centering
    \begin{minipage}[b]{0.73\textwidth}
        \centering
        \includegraphics[width=\textwidth, trim={1pt 0pt 74pt 0pt}, clip]{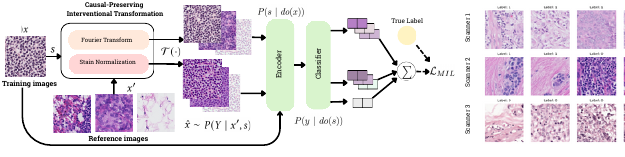}
        \label{fig:image1}
    \end{minipage}
    \begin{minipage}[b]{0.25\textwidth}
        \centering
        \includegraphics[width=\textwidth] {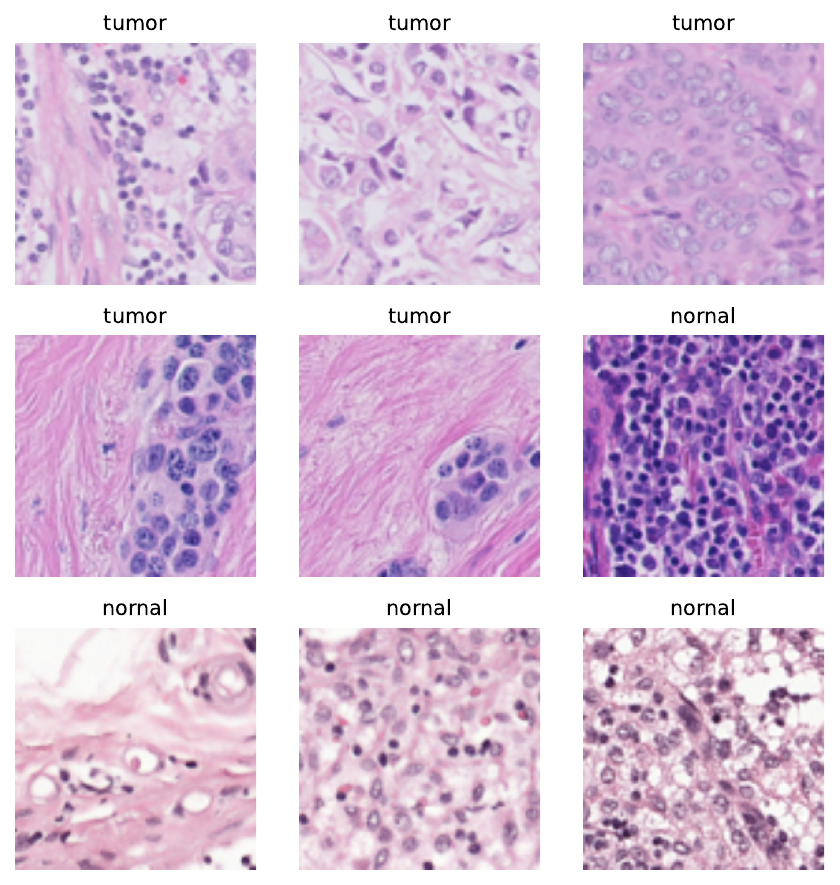}
        \label{fig:image2}
    \end{minipage}
    \caption{{Overall architecture of \mName{} (\textbf{left}) and qualitative results across three scanners on the CAMELYON17 dataset (\textbf{right}). Correctly predicted examples are shown in comparison with the baseline.}}
     \label{fig:model_structure}
\end{figure*}

\begin{enumerate}
    \item We introduce a unified causal learning framework designed to tackle domain shift in histopathology analysis by uncovering the causal relationship between histopathology images and diagnostic labels, while mitigating confounding bias. To the best of our knowledge, this is the first work that enables the training that leverages principles of causal inference to enhance out-of-distribution generalization for the histopathology analysis task. 
    \item Our design leverages the front-door adjustment principle, which does not rely on the assumption of observed confounders. We effectively implement through novel Causal-Preserving Interventional Transformation (CPIT) module that integrate semantic representations with visual instance-level observations. It simulates the effect of interventions on the visual distribution, enabling front-door identification using only observational data. 
    \item Extensive experiments on the CAMELYON17 dataset and the BMHist dataset demonstrate that the proposed method consistently outperforms purely statistical models, highlighting the power of causal reasoning to move beyond correlation and capture clinically meaningful, generalizable patterns in histopathological analysis.

\end{enumerate}

The remainder of this paper is organized as follows: Section~\ref{sec:method} describes the proposed approach in detail. Section~\ref{sec:experiments} presents the experimental results. Section~\ref{sec:conclusion} concludes the paper with potential directions for future research.

\section{Methodology}
\label{sec:method}
\subsection{Causal learning framework}
The degradation in performance on out-of-distribution data is essentially caused by the confounder which makes variable X and Y correlated even if $X$ and $Y$ have no direct causation. The confounder C draws two causal links: $C \rightarrow X$ and $C \rightarrow Y$ (Figure~\ref{fig:causal-graph} (b)). The observational distribution, therefore, can be expressed as:
\begin{equation}
    P(Y|X = x) = \sum_c P(Y|x, c)P(c| x)
    \label{eq:observervational-probability},
\end{equation}
where $c$ denotes a specific value of $C$. This represents the spurious path $X \leftarrow C \rightarrow Y$, meaning the model may rely on shortcut signals from C (e.g., scanner type or selection bias) instead of learning the true causal features in X.
Given the causal graph in Figure~\ref{fig:causal-graph}(b), the causal dependencies can learn from a mediator $S$ on the directed causal path from $X$ to $Y$. We can treat the dependencies as two parts: a semantic extractor $S$ from $X$ ($X \rightarrow S$) and a predictor from $S$ to $Y$ ($Y \rightarrow S$). The conditional distribution $P(Y|X)$ can present through $S$ as:
\begin{equation}
    P(Y |X) = \sum_s P(Y|S = s) P(S = s|X)
    \label{eq:observational-probability-with-s}
.\end{equation}
Since there is no backdoor path from $X$ to $Z$, the effect of $X$ on $S$ is identifiable via intervention probability $P(S| do(X = x))$ where the notation $do(X=x)$ or $do(x)$ denotes the intervention to X by setting its value to x \cite{pearl2016causal}, so: 
\begin{equation}
    P(S=s|do(X = x)) = P(S = s| X = x). 
\end{equation}
However, in Eq.~(\ref{eq:observational-probability-with-s}), a spurious correlation between $S$ and $Y$ may arise through the path $S \leftarrow X \rightarrow C \rightarrow Y$, which can result in biased causal estimates of $P(Y|S)$. To eliminate this spurious correlation, we block the back-door path by conditioning on $X$. The causal effect of $S$ on $Y$ becomes identifiable and can be estimated via the interventional distribution
\begin{equation}
    P(Y|do(S = s)) = \sum_{x'} P(Y|S = s, X = x')P(X=x'),
    \label{eq:p(y|do(s))}
\end{equation}
where $x'$ denote specific value of $X$.
By chaining the two partial effects, we can obtain the overall causal effect of $X$ on $Y$ using the front-door adjustment formula with intervention probability $P(Y| do(X = x))$, hence:
\begin{equation}
    \begin{split} 
        P(Y \mid do(X = x)) &= \sum_s P(S=s|do(x)) P(Y|do(S=s))\\
        &= \sum_sP(s \mid x) \sum_{x'} P(Y \mid x', s)  P(x') \\
        &= \mathbb{E}_{P(s\mid x)}\mathbb{E}_{P(x')}[P(Y \mid s, x')].
    \end{split}
\end{equation}
Here, the causal estimation no longer relies on the confounder $C$. Specifically, $P(S \mid X)$ can be implemented as an encoder as it implicitly estimates a representation $S$, while the predictor $P(Y \mid do(s))$, defined in Eq.~(\ref{eq:p(y|do(s))}), approximates an interventional process by sampling across other instances $x'$.
  

\subsection{Causal-Preserving Interventional Transformation}
The key aspect is parameterizing the predictive distribution $P(Y \mid s, x')$, where $s$ is semantic information from  the query image $x$ and $x'$ spans the broader representation space of visual styles. The more semantically meaningful information that $P(Y|s,x')$ can capture from X, the better our method can approximate $P(Y|do(x))$. As direct “physical interventions” would require passing each fixed cohort of clinical features through all possible acquisition pipelines (e.g., multiple scanners), which is impractical in histopathology, we instead introduce Causal-Preserving Interventional Transformation (CPIT) to emulate such variations. By combining Fourier-based transforms (Sec.~\ref{sub:fourier_transform}) to capture texture and contrast changes with Stain Normalization (Sec.~\ref{sub:stain_norm}) to simulate color and scanner-related shifts, CPIT generates transformed images $\{\mathcal{T}_{x'}(x)\}_{x'}$, that enables closer approximation of $P(Y \mid do(x))$ through marginalization over realistic style variations. Therefore, we can preserve the causal content $s$ of the input $x$ while incorporating non-causal variations from other instances $x'$. We therefore have:
\begin{equation}
    P(Y \mid x', s) = P(Y \mid \mathcal{T}_{x'}(x)),
\end{equation}
where $x'$ is uniformly sampled from the training data across domains to ensure coverage of scanner/stain variations.

\subsubsection{ Fourier  Transform}
\label{sub:fourier_transform}
The Fourier Transformation $\mathcal{F}$ of an image $x$ can be written:
\begin{equation}
    \mathcal{F} (x)= \mathcal{A}(x) \times e^{-j \times \mathcal{P}(x)},
\end{equation}
where $\mathcal{A}$ and $\mathcal{P}$ denote the amplitude and phase of $x$, respectively. Early studies \cite{oppenheim1979phase}, \cite{oppenheim1981importance}, \cite{piotrowski1982demonstration} have shown that the phase component retains most of the high-level semantics in the original signals, while the amplitude component majorly contains low-level statistics. Following \cite{causalstyletransfer}, we keep the phase component as content features $s$, the amplitude of original image $x$ is mixed with another style image $x'$ as follows:
\begin{equation}
    \mathcal{T}_{x'}^\mathcal{F}(x) = \mathcal{F}^{-1}((1 - \lambda)\mathcal{A}(x) + \lambda \mathcal{A}(x') \times e^{-j \times \mathcal{P}(x)}),
\end{equation}
where $\lambda \sim U(0, \eta)$ and $0 \le \eta \le 1$ is a hyperparameter controlling the maximum style mixing rate. 
\subsubsection{Stain Normalization}
\label{sub:stain_norm}
Stain normalization is a transformation function $\mathcal{T^S}$ that simulates color and scanner-related level interventions. Specifically, the goal of $\mathcal{T^S}$ is to map the color distribution of a source image $\theta_x$ to that of a reference image $\theta_{x'}$, while preserving semantic content $s$ from source image \cite{tellez2019quantifying} as:
\begin{equation}
    \theta_{x} \xrightarrow{\mathcal{T}_{x'}^S (x)} \theta_{x'}.
\end{equation}
To instantiate $\mathcal{T^S}$, we utilize the patch-based color normalization method of Reinhard et al.~\cite{reinhard2001color}, which operates in the $l\alpha\beta$ color space by aligning the mean and standard deviation of a source patch’s color distribution to match that of a reference patch. 
While we adopt Reinhard normalization, the proposed framework is not limited to this choice. Other normalization techniques, such as Macenko~\cite{macenko2009method}, Vahadane~\cite{vahadane2016structure}, or learned transformations can also serve as valid instantiations of $\mathcal{T^S}$, provided they maintain semantic consistency and introduce plausible stylistic variation across domains.

\subsection{Optimization}
We aim to approximate the interventional distribution 
$P(Y \mid do(x)) \approx \mathbb E_{p(s|x)} \mathbb E_{p(x')}$.  This leads to the training loss:
\begin{equation}
\mathcal{L}_{MIL}(x,y) 
= \mathcal{L}_{cls}\!\left(
\frac{1}{N}\sum_{i=1}^{N} F^{mix}_i, \; y
\right),
\end{equation} where the interventional expectation is approximated using averaging aggregation over $N$ transformed instances, each $F^{mix}_i$ is defined as:
\begin{equation}
F^{mix}_i 
= \gamma F(\mathcal{T}^F_{x'_i}(x)) 
+ (1-\gamma)F(\mathcal{T}^S_{x'_i}(x)),
\end{equation}
with $\gamma \in [0,1]$ controlling the weighting between Fourier-based and Stain-based transformations. To stabilize training under large inter-sample variability,  we add a residual contribution from the original input to produce the final loss:
\begin{equation}
\mathcal{L}_{MIL}(x,y) 
= \mathcal{L}_{cls}\!\left(
\beta F(x) + \frac{1-\beta}{N}\sum_{i=1}^N F^{mix}_i, \; y
\right),
\label{eq:mixed}
\end{equation}
where $0 \le \beta < 1$ balances original predictions and 
transformed predictions. 

\begin{figure*}[ht]
    \centering
    \includegraphics[width=1\linewidth]{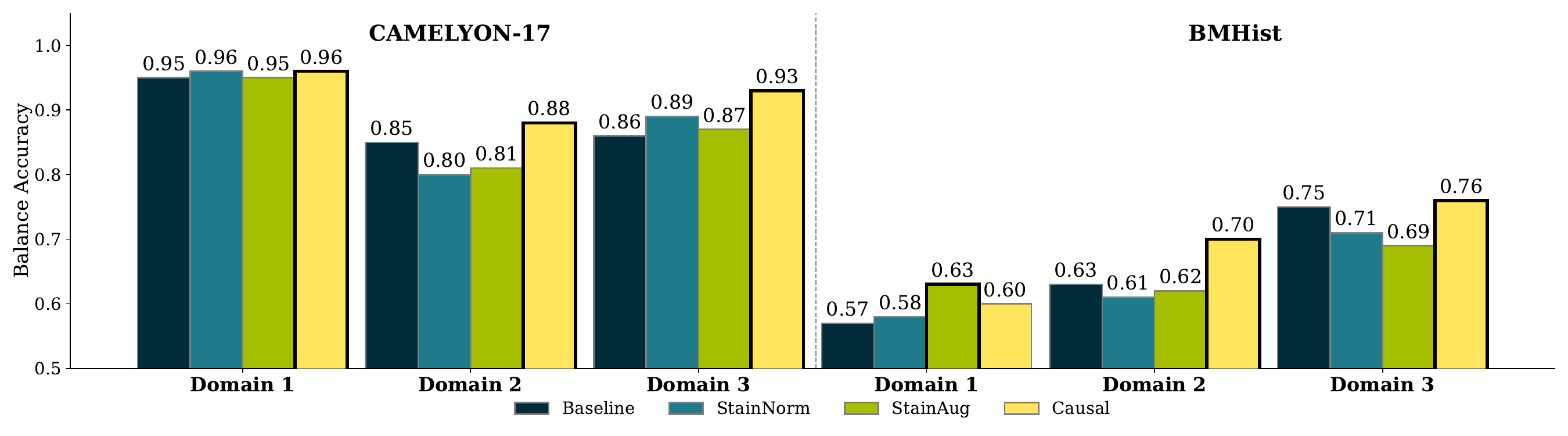}
    \caption{Leave-one-domain-out classification balance accuracies (in \%) on the CAMELYON17 and the BMHIST.}
    \label{fig:experimental-results}
\end{figure*}
\section{Experiment}
\label{sec:experiments}
We evaluate tumor classification on the CAMELYON17 dataset \cite{camelyon} and a private histopathology dataset (BMHist). In CAMELYON17, different scanners define the domains, while in BMHist the domains are manually defined by stain variations. We adopt a leave-one-domain-out setup: training on two domains and testing on the held-out domain. Results are averaged over three runs with different seeds. Following \cite{stacke2020measuring}, we report Balanced Accuracy to assess generalization under domain shift. Data are split into training/validation/testing (85/15/5) consistently across domains.
Our causal-learning method integrates Fourier Transform of weight 0.25, and Stain Normalization of weight 0.75, predictions, we choose $\beta$ value 0,2.
\subsection{Datasets}

We evaluated our proposed approach on two datasets: \textbf{CAMELYON17} \cite{camelyon} and \textbf{BMHist}.
The \textbf{CAMELYON17} dataset consists of Hematoxylin and Eosin (H\&E)-stained whole-slide images of breast lymph nodes, collected from three different scanners. For each scanner, we randomly selected slides from three patients and extracted image patches of size 500$\times$500 pixels, yielding a total of 15,000 patches. 
The \textbf{BMHist} dataset is a private collection of lung tumor slides from Bach Mai Hostpital. The dataset is part of a national research project, and its use has been approved by the institutional ethics committee. We extracted 256$\times$256 patches from slides of 44 patients and manually grouped them into three distinct stain styles, resulting in over 16,000 image patches in total.
\subsection{Experimental Results}
\begin{table}[t!]
\centering
\small
\setlength{\tabcolsep}{3pt}
\renewcommand{\arraystretch}{1.1}
\begin{tabular}{|l|c|c|c|c|}
\hline
 \textbf{Method} & \textbf{Domain 1} & \textbf{Domain 2} & \textbf{Domain 3} \\
\hline
Baseline & 0.95 & 0.85 & 0.86 \\
\mName - Stain   & 0.95 & 0.83 & 0.89 \\
\mName - Fourier   & 0.96 & 0.83 & 0.92 \\
\mName $(\gamma = 0.25)$ & \textbf{0.96} & \textbf{0.88} & \underline{0.93} \\
\mName $(\gamma = 0.5)$ & {0.96} & {0.86} & \textbf{0.94} \\
\mName $(\gamma = 0.75)$ & {0.96} & {0.85} & {0.92} \\
\hline
\end{tabular}
\caption{Performance of \mName\ under different CPIT configurations on the CAMELYON17, including removal and weighted combinations of Stain and Fourier transformations.}
\label{tab:multiple_transform_methods}
\end{table}
Following~\cite{jahanifar2025domain}, we use a pre-trained ResNet50~\cite{he2016deep} backbone and compare with Stain Normalization (StainNorm) and Stain Augmentation (StainAug). StainNorm adopts Reinhard normalization~\cite{reinhard2001color}, while StainAug applies HED augmentation~\cite{tellez2019quantifying} which achieved top performance in  domain generalization benchmarks \cite{jahanifar2025domain}.  Figure~\ref{fig:experimental-results} summarizes performances on three cross-domain where our proposed causal learning approach significantly outperforms the baseline and peer methods across all target domains.
Specifically, on the CAMELYON17 dataset, \mName outperforms the baseline by 1\% and 3\% on Domain 1 and Domain 2 and achieves a remarkable 7\% improvement  on Domain 3, exceeding the StainNorm method and StainAug by 4\% and 6\%, respectively. Qualitative results are shown in Figure~\ref{fig:model_structure} (right). The results on the BMHist dataset in Figure~\ref{fig:experimental-results} further demonstrate the superiority of our method. While StainAug performs slightly better on Domain 1, \mName achieves consistent improvements across all three domains, indicating stronger generalization and stability under domain shifts.
\subsection{Ablation Studies}
In Table~\ref{tab:multiple_transform_methods}, we show the effectiveness of the model integrating both Fourier and stain-based transformations achieves the best overall performance across all domains. We hypothesize that the superior performance of the mixed model arises from their positive effects - Fourier transforms capture texture and contrast variation, while stain normalization accounts for color and scanner discrepancies. Increasing the coefficient of the stain transformation ($\gamma = 0.25$) leads to better performance, indicating that the stain normalization better captures semantic features, while the Fourier transforms mostly models non-causal appearance. Therefore, emphasizing the stain representation yields more robust predictions.
\section{Conclusion}
\label{sec:conclusion}
This paper proposes CLEAR, a causal framework for addressing domain shift in histopathology via Causal-Preserving Interventional Transformations. By exploiting causal relationships in semantic features, CLEAR achieves consistent improvements on the CAMELYON17 and BMHist datasets, demonstrating the effectiveness of causal learning for robust clinical models. Future work will validate this approach on larger cohorts, analyze spatial context effects and general domain generalization methods, and apply it to a broader range of medical imaging tasks.
\section{Acknowledgement}
Truong Thi Kieu Anh was funded by the Master, PhD Scholarship Programme of Vingroup Innovation Foundation (VINIF), code VINIF.2023.ThS.009. We would like to express our sincere gratitude for the support and companionship of the National Foundation for Science and Technology Development (NAFOSTED) in the research project IZVSZ2\_229539, implemented from 2025 to 2027. The BMHist data is provided by Bach Mai Hostpital under the KC4.0 project: “Research on the application of artifcial intelligence in lung cancer diagnosis through analysis of chest CT images, fexible bronchoscopy images and histopathology images”, No. KC-4.0-40/19-25.", funded by the Ministry of Science and Technology of Vietnam
\vspace{12pt}

\bibliographystyle{IEEEbib}
\bibliography{references} 
\end{document}